# On Measuring the Impact of Human Actions in the Machine Learning of a Board Game's Playing Policies

Dimitris Kalles

Hellenic Open University, Sachtouri 23, 26222, Patras, Greece

kalles@eap.gr

**Abstract.** We investigate systematically the impact of human intervention in the training of computer players in a strategy board game. In that game, computer players utilise reinforcement learning with neural networks for evolving their playing strategies and demonstrate a slow learning speed. Human intervention can significantly enhance learning performance, but carrying it out systematically seems to be more of a problem of an integrated game development environment as opposed to automatic evolutionary learning.

# 1  Introduction

Several machine learning concepts have been tested in game domains, since strategic games offer ample opportunities to automatically explore, develop and test winning strategies. The most widely publicised results occurred during the 1990s, when IBM made strenuous efforts to develop (first with Deep Thought, later with Deep Blue) a chess program comparable to the best human player.

As early as 1950, Shannon (1950) studied how computers could play chess and proposed the idea of using a value function to compete with human players. Following that, Samuel (1959) created a checkers program that tried to find "the highest point in multidimensional scoring space", only to have his research rediscovered by Sutton (1988) who formulated the TD($\lambda$) method for temporal difference reinforcement learning (RL). Since then, more games such as Tetris, Blackjack, Othello (Leouski, 1995), chess (Thrun, 1995) and backgammon (Tesauro, 1992; 1995) were analysed by applying TD($\lambda$) to improve their performance.

TD-Gammon (Tesauro, 1992; 1995) was the most successful early application of TD($\lambda$) for the game of backgammon. Using RL techniques and after training with 1.5 million self-playing games, a performance comparable to that demonstrated by backgammon world champions was achieved.

As far as strategy games are concerned, the most important and critical point of them is to select and implement the computer's strategy during the game. The term *strategy* stands for the selection of the computer's next move considering its current situation, the opponent's situation, consequences of that move and possible next moves of the opponent. RL helps solve this problem by formulating strategies in terms of policies. In theory, the advantage of RL to other learning methods is that the target system itself detects which actions to take via trial and error, with limited need for direct human involvement.

In our research, we use a new strategy game to gain insight into the question of how game playing capabilities can be efficiently and automatically evolved. The problem that we aim to highlight in this paper is that the usual arsenal of computational techniques does not readily suffice to develop a winning policy and that one must couple automation with careful experimental design. Our contribution is the development and experimental validation of simple quantitative indices, that measure performance improvement of automatic game playing, to better support our decisions about which training paths to follow. For this reason we have designed and carried out several experimental sessions comprising in total well over 400,000 computer-vs.-computer games and over 200 human-vs.-computer games.

The rest of this paper is organised in four sections. The next section presents the basic details of the game and its introductory analysis. The third section describes our experimentation on training. The fourth section discusses the impact and the limitations of human-assisted learning and states the recommended directions for future development. The concluding section summarises the work.

## 2  The Game Description

The game is played on a square board of size *n,* by two players. Two square bases of size *a* are located on opposite board corners. The lower left base belongs to the white player and the upper right base belongs to the black player. At game kick-off each player possesses *β* pawns. The goal is to move a pawn into the opponent's base.

The base is considered as a single square, therefore every pawn of the base can move at one step to any of the adjacent to the base free squares (see Fig. 1 for examples and counterexamples of moves). A pawn can move to an empty square that is vertically or horizontally adjacent, provided that the maximum distance from its base is not decreased (so, backward moves are not allowed). Note that the distance from the base is measured as the maximum of the horizontal and the vertical distance from the base (and not as a sum of these quantities). A pawn that cannot move is lost (more than one pawn may be lost in one round). If some player runs out of pawns he loses.

The leftmost board in Fig. 1 demonstrates a legal and an illegal move (for the pawn pointed to by the arrow). The rightmost boards demonstrate the loss of pawns (with arrows showing pawn casualties). Such (loss incurring) moves bring about the direct adjustment of the moving pawn with some pawn of the opponent. In such cases the "trapped" pawn automatically draws away from the game. As a by-product of this rule, when there is no free square next to the base, the rest of the pawns of the base are lost.

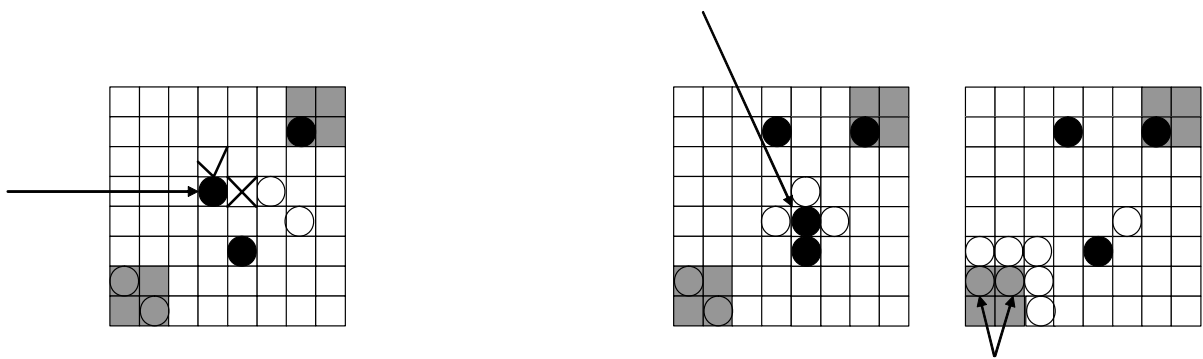

**Fig. 1.** Examples and counterexamples of moves

## 2.1 The Analysis Context

Past research (Kalles & Kanellopoulos, 2001) initially demonstrated that, when trained with self-playing games, both players had nearly equal opportunities to win and neither player enjoyed a pole position advantage. Follow-up research (Kalles & Ntoutsi, 2002) furnished preliminary results that suggested a computer playing against itself would achieve weaker performance when compared to a computer playing against a human player.

The game is a discrete Markov procedure, since there are finite states and moves. Each episode terminates and the game is thus amenable to analysis by reinforcement learning (Sutton & Barto, 1998). The *a priori* knowledge of the system consists of the rules only. The agent's goal is to learn a policy that will maximize the expected sum of rewards in a specific time; this is called an optimal policy. A policy determines which action should be taken next given the current state of the environment. As usual, at each move the agent balances between choosing an action that will straightforward maximize its reward or choosing an action that might prove to be better. A commonly used starting $\varepsilon$-greedy policy with $\varepsilon=0.9$ was adopted, i.e. the system chooses the best-valued action with a probability of 0.9 and a random action with a probability of 0.1.

At the beginning all states have the same value except for the final states. After each move the values are updated through TD($\lambda$), where $\lambda$ determines the reduction degree of assigning credit to some action and was set to $\lambda=0.5$.

Neural networks were used to interpolate between game board situations (one for each player, because each player has a unique state space). The input layer nodes are the board positions for the next possible move, totalling $n^2-2a^2+10$. The hidden layer consists of half as many hidden nodes, whereas the output node has only one node, which can be regarded as the probability of winning when one starts from a specific game-board configuration and then makes a specific move.

Note that, drawing on the above and the game description, we can conclude that we cannot effectively learn a deterministic optimal policy. Such a policy does exist for the game (Littman, 1994), however the use of an approximation (neural network) effectively rules out such learning. Of course, even if that was not the case, it does not follow that converging to such a policy is computationally tractable (Condon, 1992).

## 3 The Experimental Setup

To focus on how one could measurably detect improvement in automatic game playing, we devised a set of experiments with different options in the reward policies while at the same time adopting a relatively narrow focus on the moves of the human (training player).

Herein we report experiments with 8 game batches. Each game batch consists of 50,000 computer-vs.-computer (CC) games, carried out in 5 stages of 10,000 games each. For batches that have involved human-vs.-computer (HC) games, each CC stage is interleaved with a HC stage of 10 games. Thus, HC batches are 50,050 games long. In HC games, a human is always the white player.

We now show the alternatives for the human player in Table 1. Briefly describing them, we always move from the bottom-left base to the north and then move right, attempting to enter the black base from its vertical edge (see Fig. 2). We explore a mix of learning scenarios, whereby at some experiments we explicitly wander around with the human player, allowing the black player to discover a winning path to the white base. Of course, if the human player wants to win, this is straightforward.

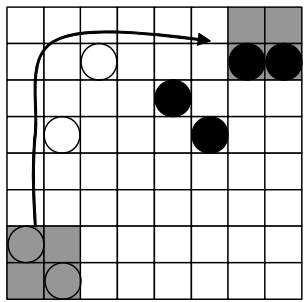

**Fig. 2.** The path of a human player

**Table 1. Policies of white human player**

| | White player always starts from bottom-left |
|---|---|
| 1 | North, then Right, attempting to enter from vertical edge |
| 2 | Allows Black to win 5 games, then Policy #1 |
| 3 | Allows Black to win 10 games |

The alternatives for the rewards are shown in Table 2. Briefly describing them, the first reward type (as documented in [8]) assigns some credit to states that involve a pawn directly neighbouring the enemy base. It also rewards captured pawns by calculating the difference of pawn counts and by scaling that difference to a number in [-1,1]. The other two polices have been developed with a view towards simplification (by dropping the base adjacency credit) and towards better value alignment.

**Table 2. Reward types**

|   | White player always starts from bottom-left |
|---|---|
| 1 | Win: 100, Loss: -100<br>Win-at-next-move: 2, Loss -at-next-move: -2<br>Pawn difference scaled in [-1,1] |
| 2 | Win: 100, Loss: -100<br>Pawn difference scaled in [-1,1] |
| 3 | Win: 100, Loss: -100<br>Pawn difference scaled in [-100,100] |

### 3.1 Varying only the white player policy (HC)

The white player policy can be deliberately varied in HC games only, of course. We report below the results of three HC batches, where the reward type was set to 1. A short description of the batches is shown in Table 3, whereas the results are shown in Fig. 3, Fig. 4 and Fig. 5.

**Table 3. Description of batches 1 - 3**

|   | Game Type – Reward - Policy |
|---|---|
| 1 | HC, 1, 1 |
| 2 | HC, 1, 2 |
| 3 | HC, 1, 3 |

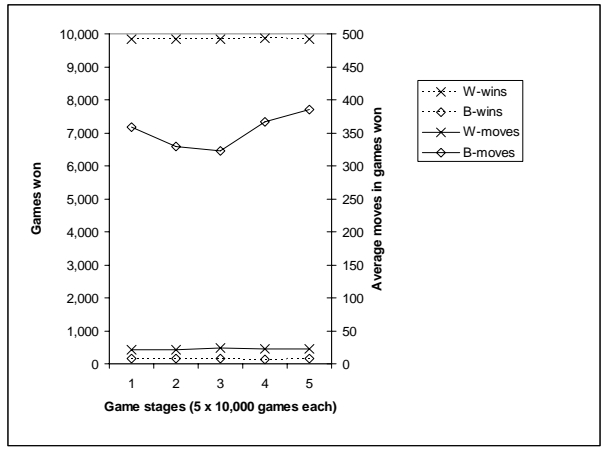

**Fig. 3.** 1st Experimental Session.

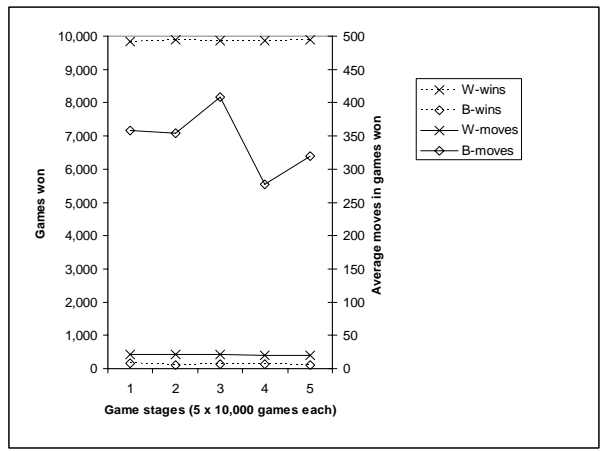

**Fig. 4.** 2nd Experimental Session.

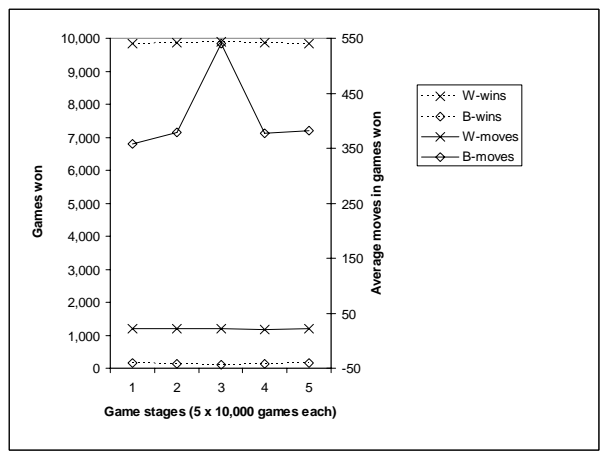

**Fig. 5.** 3rd Experimental Session.

Page 8 of 19

First, we observe that the white player overwhelms the black one. The black player's performance can be seen to improve only in the third batch, where the white player aims not to win. Still, this improvement is indirect (note that the white player requires a larger average number of moves to win). This suggests that the number of CC games may be too few to allow the pole position advantage of the white player to diminish. In that respect, it also seems that when the white player wins, even in few of the games, the efficiency of the human-induced state-space exploration can be picked up and sustained by the subsequent exploitation of the CC stage.

### 3.2 Varying only the reward (HC)

In the next experimental round, we froze the policy type to 1. A short description of the batches is shown in Table 4, whereas the results are shown in Fig. 3 (earlier page), Fig. 6 and Fig. 7.

**Table 4.** Description of batches 1, 4, 5

|   | Game Type – Reward - Policy |
|---|---|
| 1 | HC, 1, 1 |
| 4 | HC, 2, 1 |
| 5 | HC, 3, 1 |

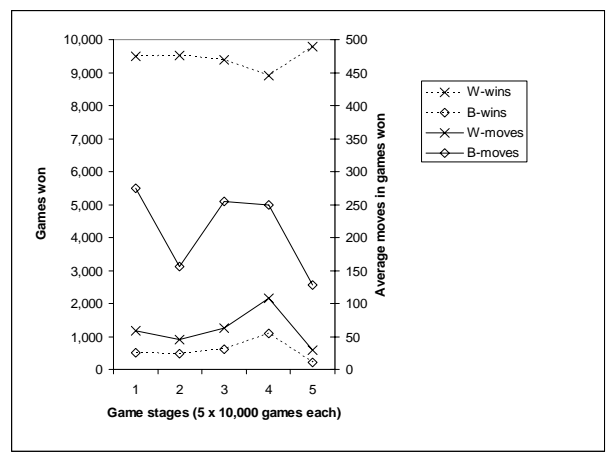

**Fig. 6.** 4th Experimental Session.

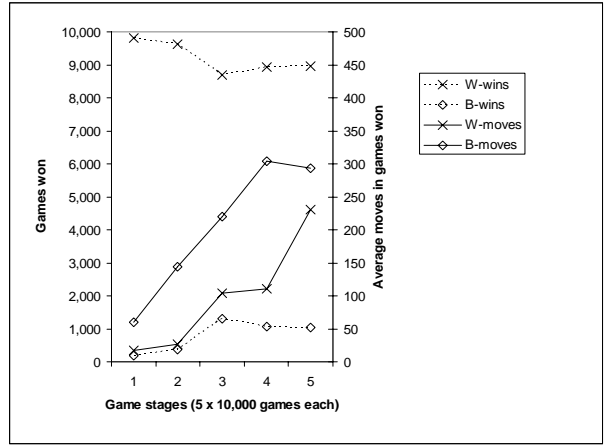

**Fig. 7.** 5[th] Experimental Session.

We now observe that the fifth experiment suggests a clear improvement for the black player, as the number of moves required by the white CC player consistently increases. In any case, we also observe again that as the while HC player wins, so does the white CC player seem able to sustain its own wins. Note also the highly irregular behaviour in the fourth batch, where the fourth stage witnesses a strong turn. It is interesting that this is associated with a superficially rewarded pawn advantage. These results are an indication that pawn advantage rewards should be commensurate with their expected impact in the game outcome; losing many pawns and not capturing any indicates that we are about to lose.

### 3.3 Interleaving CC and HC games

The above findings lead us to experiment with the following scenario: first conduct a CC batch, then use its evolved neural network to spawn experimentation in two separate batches, one CC and one HC. Note that, until now, all experiments were *tabula-rasa* in the sense that each experiment started with a "clean" neural network. In all these experiments, we used the reward type (3) that aligns pawn advantage with winning a game. A short description of the batches is shown in Table 5, whereas the results are shown in Fig. 8, Fig. 9 and Fig. 10.

**Table 5.** Description of batches 6 - 8

|   | Game Type – Reward - Policy |
|---|---|
| 6 | CC, 3, - |
| 7 | CC, 3, - (based on batch 6) |
| 8 | HC, 3, 1 (based on batch 6) |

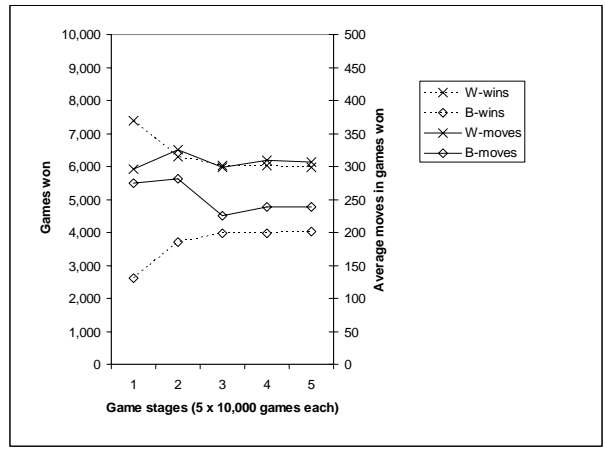

**Fig. 8.** 6[th] Experimental Session.

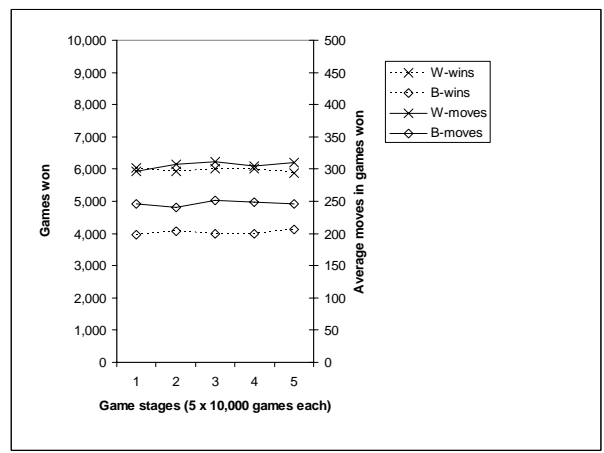

**Fig. 9.** 7[th] Experimental Session.

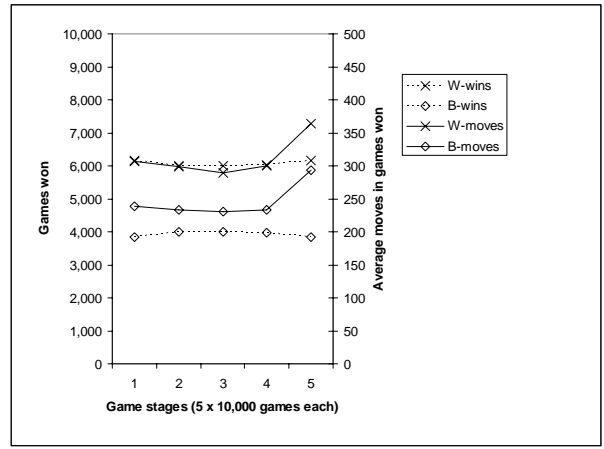

**Fig. 10.** 8<sup>th</sup> Experimental Session.

The succession of the two CC games is quite revealing: while the white player does enjoy a pole position advantage, this is subsequently eroded. This is surely obvious if one observes the number of games won. Note that the short distance between the lines showing the average number of moves is another testimony. This further supports our initial experimentation that suggested both players had equal chances to win; our experiments now show that this equality is progressively brought about by the convergent behaviour of the two players.

Quite as importantly, one can note that the introduction of human intelligence eventually allows the white CC player to immediately increase the performance gap.

It is most instructive to compare the fifth and eighth batches (Fig. 7 and Fig. 10 respectively) since their only difference is that the eighth batch is based on a previous CC batch. It seems that the CC batch has instilled some knowledge in the white player that stabilizes its behaviour relative to the black player. This very inefficient knowledge is not straightforward to "forget" in order to be replaced by human playing RL values. Perhaps, it would be more precise to call it knowledge inertia since, if left unattended to, the two computer players will most likely reach an uninteresting equilibrium (as Fig. 8 and Fig. 9 show).

### 3.4 Measuring human impact

There is a relatively straightforward way to judge when the human intervention had its "best" impact: we can pit against each others the players of the fifth and eight batches. In essence, we measure whether the same amount of human training was more effective in delivering a better computer player when starting from *tabula rasa* or when starting from some automatically generated knowledge (note that the eighth batch was built on twice as many CC games compared to the fifth batch). This was done in two steps (of 1,000 CC games each) where, in the first step the white player of the fifth batch plays again the black player of the eighth batch (in the second step a similar setting applies). The results are shown in Table 6.

**Table 6.** Comparative evaluation of distinct learning paths

|         | Games Won |       | Average # of Moves |       |
|---------|-----------|-------|--------------------|-------|
|         | White     | Black | White              | Black |
| W5 – B8 | 715       | 285   | 291                | 397   |
| W8 – B5 | 530       | 470   | 445                | 314   |

From this viewpoint, the difference is striking: the fifth batch seems to have generated far more effective players, both white and black. While the white player seems to win easily in the W5-B8 round, the black player seems to give its white opponent a hard time in the W8-B5 round. This reinforces our earlier observation about the knowledge inertia of the eighth batch.

What we cannot yet tell is what amount of this performance is directly due to the white player being trained by the human player and what amount flows to the black player who is opposed to a good white player.

To attempt to answer this question we have repeated the experiments of batches 5 through 8 in accelerated learning mode, in which a CC batch is now carried out in 5 stages of 1,000 games each. The new results are shown in Table 7.

**Table 7.** Comparative evaluation of distinct learning paths in accelerated learning mode

|             | Games Won |       | Average # of Moves |       |
|-------------|-----------|-------|--------------------|-------|
|             | White     | Black | White              | Black |
| W5* – B8*   | 426       | 574   | 258                | 176   |
| W8* – B5*   | 109       | 891   | 337                | 68    |

Again, the (modified) fifth batch is better. But now, it is also easy to observe the decrease of the performance of the white player and the decrease in the average number of moves that it takes the black player to win. Both are substantial and suggest that we have managed to train the black player far more effectively than the white one!

In fact, it seems that the accelerated learning mode may be more warranted than the initial one. Shorter CC game batches may limit the wandering into unexplored paths with limited eventual credit;

also, they do not easily dilute the experience generated by human involvement. To further support this claim we present Table 8 below, which summarizes the results for exhaustive cross-testing.

**Table 8.** Comparative evaluation of distinct learning paths in accelerated learning mode (cont'd)

|   |   | Games Won | | Average # of Moves | |
|---|---|---|---|---|---|
|   |   | White | Black | White | Black |
| A. | W5 – B5* | 578 | 422 | 316 | 409 |
|    | W5* – B5 | 642 | 358 | 298 | 299 |
| B. | W5 – B8* | 660 | 340 | 273 | 214 |
|    | W8* – B5 | 651 | 349 | 293 | 299 |
| C. | W8 – B5* | 67  | 933 | 269 | 47  |
|    | W5* – B8 | 387 | 613 | 386 | 221 |
| D. | W8 – B8* | 61  | 939 | 166 | 22  |
|    | W8* – B8 | 530 | 470 | 388 | 258 |

The wide performance gap in terms of games won that is evident in Table 8.C and Table 8.D has a very intuitive explanation. When a white human player defeats the black player, the backward credit updates for the black player affect a smaller range of possible moves (since, such moves revolve around the black base) and the transition probabilities for such moves are better approximated. Thus, the black player converges faster to a defending strategy. When the advantage of the human player disappears, the black player can then tramp over the white one; even more so when white training is based on long automatic-playing sessions. At first sight, this seems to confirm an earlier similar observation on the preference of Q-learning players to first develop defensive tactics (Littman, 1994).

Therein seems to be observed a recurring theme which, we believe, is a fundamental demonstration of indirect learning: give the machine a robust learning strategy and the luxury of a good opponent at well-placed checkpoints and you may expect it to improve over time. Well-placed checkpoints are not too far apart so that knowledge may be "forgotten" and not too close so that knowledge chunks "overlap". It is intriguing that these are also hallmarks of human-to-human tutoring.

## 4 Discussion

We definitely need more experiments if we are to train our computer players to a level comparable to that of a human player. The options can be numerous, but we can name a few obvious ones that are also clearly independent between them: experimentation with more exploration-exploitation trade-offs or alternative RL parameters, experimentation with the learning parameters or the input-output representation of the neural network, experimentation with alternative reward types or human playing policies. Last but not least, any combination of the above may be a plausible one to investigate. In fact, we cannot directly attribute which part of the learning inefficiencies spotted in the long experimental runs of the above section may be due to the parameters of the reinforcement learning algorithm, or the parameters of the neural network.

The results we have obtained to-date clearly suggest that it is very important to find an efficient and effective way to achieve learning. We must optimize the use of expensive resources (human player) so that they are intelligently placed at critical points of the learning process, which will mostly be done automatically. Note that even though the number of HC games is relatively very small to the number of CC games, the impact of HC games can be clearly detected. Accurately measuring this impact is not straightforward, however. Therefore, it is of less importance to discuss how much to increase human involvement as opposed to gauging how to best spread a given amount of such involvement.

The question, of course, is "which options to select for experimentation". In answering this question, there are two major directions to follow (Ghory, 2004).

The first one is to devise an experimentation engine that will attempt to calculate the best parameters for effective and efficient learning. This option has conceptual simplicity, technical appeal and has delivered some interesting results (Partalis *et al.*, 2006). However, we believe that it would be an expensive addition to an already expensive task, as, still more parameters must be specified (and experimented with). By deploying a meta-experimentation level, we practically shift the problem. Moreover, we would have to define the "supervision" level of the learning process and craft appropriate measures. Beyond the number of games won and the average number of moves in such games which seem to be good candidates for this task, we may also have to come up with measures of interestingness (van den Herik *et al.*, 2005).

The second one is to embed some ad-hoc knowledge into the learning process. This is not a new concept; a combination of RL and Explanation Based Learning was initially supposed to be able to benefit the game with faster learning and the ability to scale to large state spaces in a more structured manner (Dietterich & Flann, 1997). Why, then, did it not materialize in published benefits? We believe that this is due to the inherent difficulty of attempting to merge numeric and symbolic representation and classification paradigms, and especially so in the context of large experimentations, where the coarse or fine resolution of the merging process might result in substantially different outcomes.

In retrospect, both directions seem to suggest that the numeric approach to automatic learning has some very pronounced practicality limitations. Simply put, some domains are too premature (in how we comprehend them) to lend themselves to general-purpose evolutionary improvement. It is for this reason, we believe, that our experiments demonstrate measurable improvements when subjected to human "tutoring". Though automatic playing has long been testified to deliver good results (Tesauro, 1995; Chellapilla & Fogel, 1999) and still is a vibrant area, we emphasize human impact in a new game (simple, yet state-space consuming) because we are interested precisely in exploring *disturbance* during learning, not unlike the dice in back-gammon. Note, that an interesting and probably useful extension would be to develop a minimax (computer) player and then use that player as a teacher for the learning computer (Littman, 1994).

Interactive evolution might be promising, however. In such a course, one would ideally switch from focused human training to autonomous crawling between promising alternatives. But, as we have discovered, during the preparation of this work, the interactivity requirements of the process of improving the computer player is very tightly linked to the availability of a computer-automated environment that supports this development. Such an environment was not available and was missed. As a matter of fact, the above experiments may have cost in total about one month of computing from a relatively high-end desktop. The visualization of the intermediate results, the data processing and visualization, as well as the selection of which extract of the results to choose may have cost about twice as much in human resources. (On top of that human resources cost, it would be difficult to quantify the context switching effort between several other occupations.)

This may be sustainable if we want to provide some incremental improvement to automatic game playing but it seems hardly sustainable if we aim to develop the level of automatic playing to that of the human player. In terms of the experiments described above, we have noticed several features of an experimentation system that we have deemed indispensable if one views the project from the point of system efficiency. Such features range from being able to easily design an experimentation batch, direct its results to a specially designed database (to also facilitate reproducibility), automatically process the game statistics and observe correlations, link experimentation batches in terms of succession, while at the same time being able to pre-design a whole series of linked experiments with varying parameters of duration and succession and then guide the human player to play a game according to that design.

As it seems, being able to provide a tool that captures the lifecycle of the development of an AI application is a strong contributor to the success of the take-up of that application. Perhaps, it is not surprising that when data mining (which, in principle, is close to what this research is about) started its applied steps, it was with the availability of workflow-like tools such as Clementine

(http://www.spss.com/clementine) that researchers and practitioners alike managed to navigate efficiently through the data mining process. In that sense, we aim to pursue these directions towards the automatic discovery of knowledge in game playing as opposed to equipping the computer with more detailed domain modelling (Junghanns & Schaeffer, 2001) or with standard game-tree search techniques.

# 5 Conclusion

This paper focused on the presentation of carefully designed experiments, at a large scale, to support the claim that human playing can *measurably* improve the performance of computer players in a board game.

After describing the experimental setup, we presented the results which are centred on two key statistics: number of games won and average number of moves in games won. Arguably, the high level of abstraction of these statistics should render them (as well as the proposed process) useful in the development process and evaluation of similar board games.

The computation of these statistics is a trivial task, but the key challenge is how to decide the succession of experiments, taking into account that each experiment is specified by some parameters, so as to efficiently and effectively guide the learning process. The importance of such guiding is underlined by the fact that we must thoughtfully exploit human contribution, which will undoubtedly be a scarcely available resource.

We have concluded that while most of the fundamental AI arsenal needed is already available significant applied research is required for the establishment of tools that will streamline the experimentation process. We believe that workflow-like tools will first beat the path of such streamlining before we effectively address the autonomous management of this process.

## Acknowledgements

This paper shares some setting-the-context paragraphs with two referenced papers by the same author. However, it duplicates neither methodology nor experiments nor results. All game batches have been recorded by storing the resulting neural networks, for potential examination and reproducibility tests. The results and the code are available on demand for academic research purposes.